%% file: main.tex
\ifdefined\pdfminorversion\pdfminorversion=7\fi
\documentclass{article}
\usepackage{arxiv}
\usepackage{natbib}
\usepackage[T1]{fontenc}
\usepackage{amsmath,amssymb}
\usepackage{dsfont}
\usepackage{graphicx,booktabs,longtable,array,tabularx}
\usepackage{placeins}
\usepackage{float,needspace}
\usepackage{enumitem}
\usepackage{hyperref}
\usepackage{url}

\input{preprint_layout}
\title{Beyond Benchmark Accuracy: A Diagnostic Test for LLM Chemical Prediction}
\renewcommand{\shorttitle}{Beyond Benchmark Accuracy}
\input{authors}
\hypersetup{pdftitle={Beyond Benchmark Accuracy: A Diagnostic Test for LLM Chemical Prediction}, pdfauthor={Yanyi Su, Yu Zeng, Hongshuai Wang, Chenyan Wang, Zhifeng Gao, Xiaohui Yang}, pdfsubject={Preprint}}
\begin{document}
\maketitle
\begin{abstract}
Large language models have achieved strong performance on chemical prediction benchmarks. However, existing evaluation methods typically provide the identity information of the target entity directly, making it difficult to determine whether model predictions arise from transferable structure–property reasoning capabilities or rely on target-related contextual information or memory retrieval. We propose the Identity-Blinded Reasoning Diagnostic framework to evaluate chemical prediction models when target entity identity information is hidden. In this framework, the target is no longer directly defined by its entity identity, but instead by a known source entity and an explicit structural edit operation. This setting preserves the information required to determine the target while reducing the opportunity for models to access target identity cues. We evaluate this framework across six tasks spanning odor prediction, molecular property prediction, retrosynthesis, and reaction yield prediction. The results show that models generally achieve better prediction performance when the properties of the source entity and target entity are consistent. Providing explicit source entity labels further strengthens this dependence. Moreover, while keeping the target entity, structural change, and ground-truth label unchanged, replacing only the source entity label causes some model predictions to systematically shift toward the substituted source labels.These results indicate that high benchmark accuracy in chemical prediction tasks does not necessarily imply that large language models possess reliable structure–property reasoning capabilities. Our diagnostic framework provides a controlled evaluation approach for testing whether models can genuinely update predictions according to entity changes. The framework can be extended to other AI for Science tasks, particularly those where targets can be defined through transformations of known entities.
\end{abstract}

\begin{figure}[!t]
\centering
\includegraphics[width=\linewidth]{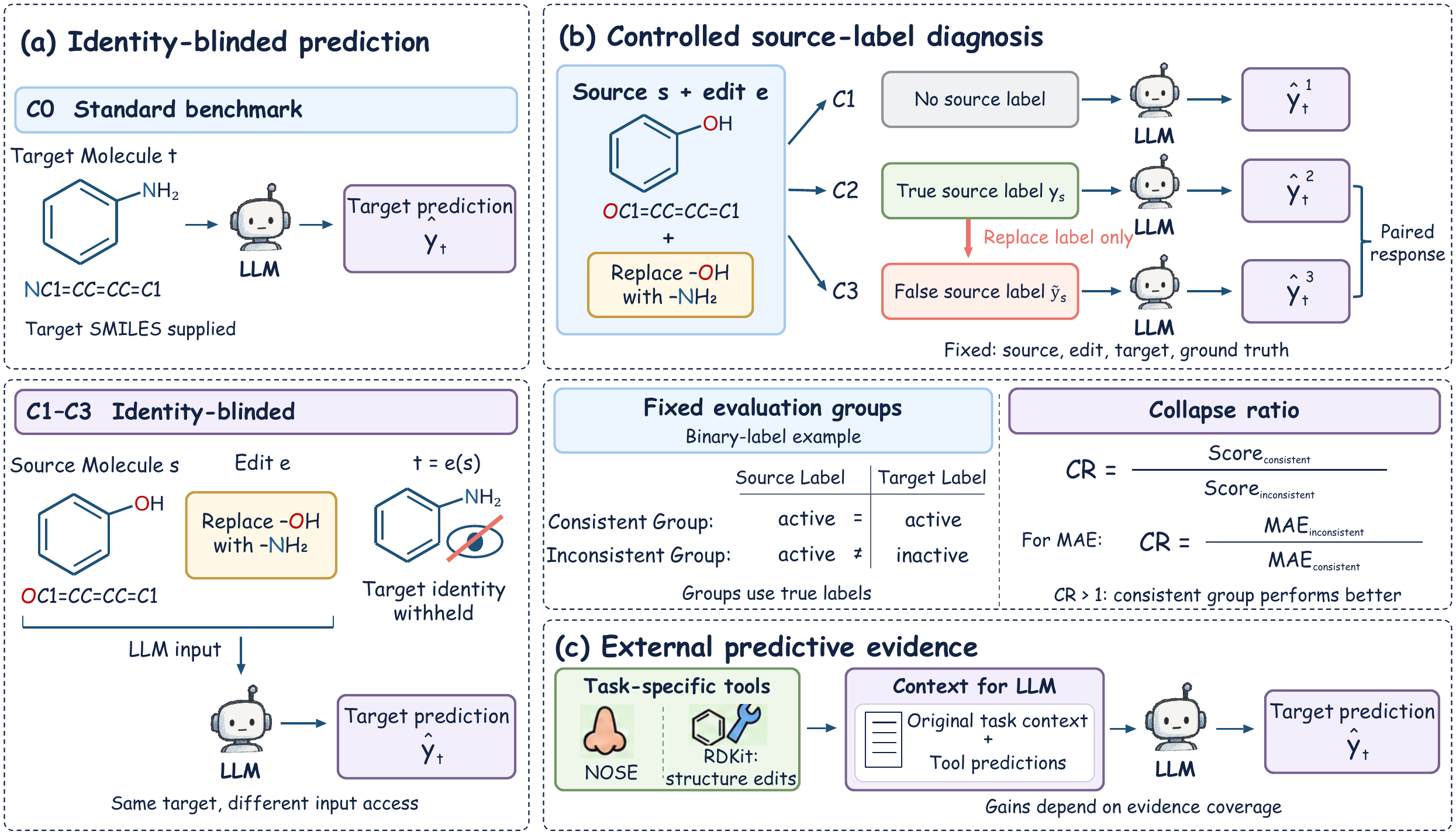}
\renewcommand{\thefigure}{1}
\caption{Overview of the Identity-Blinded Reasoning Diagnostic. (a) Comparison between standard benchmark prediction (C0) and identity-blinded prediction (C1–C3), where the target identity is hidden and defined through a source entity and edit description. (b) Source label control conditions and group analysis. C1 provides no source label, C2 provides the true source label, and C3 replaces it with a false source label. CR denotes the performance ratio between label-consistent and label-inconsistent groups. (c) Contextual evidence augmentation, where independent predictions from task-specific predictors are added as external evidence.}
\label{fig:1}
\end{figure}

\section{Introduction}

Large language models have demonstrated competitive performance on various chemical prediction tasks \citep{guo2023can}, with results in some few-shot settings matching or exceeding those of conventional methods \citep{baker2025molecular}. Chemical prediction requires models to infer properties or reaction outcomes from structure--property relationships, but whether these high scores reflect a reliable, transferable structure--property reasoning capability still requires systematic testing.

Previous studies have found that strong performance in chemical prediction may partly arise from cues other than structure--property relationships. In materials prediction, models that receive only bibliographic metadata, such as author names, publication venues, and publication years, can perform comparably to models using chemical descriptors on some tasks \citep{jablonka2026clever}. For LLMs, different representations of the same molecule may yield inconsistent predictions \citep{yan2025inconsistency}, and predictions may also directly copy labels from contextual examples \citep{zhao2023scientific}. Another study found that models perform well in molecular weight prediction, but their prediction performance declines substantially when property values undergo nonlinear transformations \citep{joe2026evaluating}.

Existing diagnostic approaches have attempted to reduce memory cues by masking context or transforming inputs, but some of these operations also alter the information on which models depend for chemical reasoning. For example, under some conditions, Busch et al. hide the meanings of chemical properties and alter the encoding of chemical structure representations through character substitutions \citep{busch2026context}. Their subsequent study notes that these substitutions also remove chemical information, making it difficult to attribute the performance decline entirely to blocked access to memorized information \citep{busch2026molecular}. A diagnostic method is therefore needed that reduces target-identity-related cues while preserving the chemical information required for structure--property reasoning.

To this end, we propose the Identity-Blinded Reasoning Diagnostic, which reduces model dependence on target-identity-related cues by withholding direct identity information, such as the target molecule's name and SMILES. As shown in Figure~\ref{fig:1}, we provide a source molecule's structure and describe in natural language the structural changes required to obtain the target molecule, asking the model to predict target properties based on the source structure and the structural changes linking it to the target. We further provide the source molecule's known property label as a reference and explicitly state that this label belongs to the source molecule. We compare model performance with and without source labels on samples whose source and target molecules have consistent or inconsistent labels. We then hold the source structure, structural change description, and prediction target fixed, replace only the true source label with a false source label, and observe how target predictions change.

We conduct a unified diagnostic across six tasks in four domains, including odor prediction, molecular property prediction, retrosynthesis, and reaction yield prediction. Independent structure reconstruction controls for the molecular tasks show that models can usually recover the target molecule's structure from the source molecule's structure and a natural-language structural change description. Without source labels, the label-consistent group already outperforms the label-inconsistent group on several tasks, and this pattern becomes more widespread when true source labels are provided, as shown in Figure~\ref{fig:3}. Replacing source labels with false source labels further causes binary, set-valued, and numerical predictions to change markedly with the substituted labels, as shown in Figure~\ref{fig:4}. Supplementing the context with task-relevant tool predictions improves performance on some tasks, but the benefits vary across tasks, as shown in Figure~\ref{fig:5}. These results reveal reliability problems in structure--property prediction by the tested LLMs and indicate that high benchmark scores are insufficient to show that models possess a reliable, transferable chemical reasoning capability.

Our main contributions are as follows:

\begin{enumerate}[leftmargin=*]
\item
We propose the Identity-Blinded Reasoning Diagnostic to evaluate the reliability of structure--property reasoning in LLM chemical prediction while preserving target structure information and property semantics.

\item
Through controlled experiments across domains, we reveal systematic dependence of the tested models' predictions on non-target cues, such as source molecule structures and labels.

\item
We conduct contextual evidence augmentation experiments to examine the improvements in prediction brought by task-relevant evidence and their relationship to tool coverage.
\end{enumerate}

\section{Related Work}

\textbf{Chemical prediction benchmarks for LLMs.} Existing benchmarks cover a range of chemical tasks, including molecular properties, reactions, retrosynthesis, yields, and molecule generation \citep{guo2023can,huang2026chemeval}. In molecular property prediction, models typically receive the target molecule's name or structure string directly. Related studies compare property prediction performance across different string representations \citep{baker2025molecular}, and odor benchmarks evaluate models using compound names and isomeric SMILES separately \citep{makri2026benchmark}. These inputs provide target structure information but may also expose target-identity-related cues, enabling models to make predictions using learned molecular knowledge or associated information. Benchmark scores alone therefore make it difficult to determine whether model performance actually arises from structure--property reasoning.

ChemIQ further examines structural understanding and property inference through manually constructed tasks. Its structure--property relationship questions provide the SMILES and scores of several molecules and ask models to predict the score of another molecule, whose SMILES is also provided directly. These scores are sums of randomly assigned substituent values rather than measured chemical properties \citep{runcie2026assessing}. This design reduces the possibility of directly recalling existing property answers, but it mainly tests whether models can learn artificially defined relationships and does not yet answer whether high performance in real chemical prediction arises from reliable structure--property reasoning. We preserve the original chemical prediction tasks and the correct answers for the targets, specify targets through source structures and structural change descriptions without directly providing their names or SMILES, and examine whether models can reliably predict from structure--property relationships under these conditions.

\textbf{Diagnosing reasoning capability in chemical prediction.} Studies have begun to diagnose models' chemical prediction capabilities through prediction consistency, structural understanding, and reasoning processes. Yan et al. and ChemDIRT compare predictions across molecular representations or instructions to examine accuracy and consistency \citep{yan2025inconsistency,inae2026chemdirt}. MolecularIQ checks structural understanding using tasks whose answers can be computed directly from molecular graphs, whereas ChemCoTBench-V2 verifies intermediate chemical steps in model outputs and identifies cases where the final answer is correct but the steps fail verification \citep{bartmann2026moleculariq,guo2026answers}.

Another line of research changes inputs or property labels to examine whether models depend on memory and simple cues. Joe et al. transform property values to examine whether models can learn new property relationships from examples \citep{joe2026evaluating}. Busch et al. hide property names, transform labels, and replace SMILES characters to study the roles of pretrained knowledge and contextual information \citep{busch2026context}. However, hiding property meanings or altering familiar chemical representations may also hinder models' use of relevant chemical knowledge. Their subsequent study explicitly notes that some performance loss after character substitutions arises because the task itself becomes harder and cannot be attributed entirely to blocked access to memorized information \citep{busch2026molecular}. Performance declines alone therefore make it difficult to distinguish whether models have lost non-target cues or have also lost information needed for chemical reasoning.

The influence of contextual labels has also been studied directly. Zhao et al. replace true property values in examples with random values while retaining the target SMILES and find that Galactica copies contextual labels, and that such copying does not always come with larger prediction errors \citep{zhao2023scientific}. This setting can examine the influence of labels but still provides the target's identity directly.

We withhold direct target names and SMILES while preserving the information needed to determine the target through the source structure and structural change description, as well as the meaning of the predicted property. We use this setting to diagnose the reliability of structure--property reasoning in real chemical prediction tasks, reducing the interference in result interpretation caused by performance changes from removing necessary chemical information.

\section{Identity-Blinded Reasoning Diagnosis}

\subsection{Target Definition and the Identity-Blinded Setting}

We specify a target entity $t=e(s)$ through a source entity $s$ and an edit description $e$, and ask the model to predict the target entity's label $y_t$. The source entity and edit jointly specify the prediction target, denoted by $t=e(s)$. The model receives only the source structure and edit information, without direct access to the target name or complete target SMILES. Edits in molecular tasks include functional group changes, fragment additions and deletions, and chirality inversions. In the reaction yield task, the target system is defined by replacing one component in the source formulation. This setting preserves the information needed to determine the target while keeping the prediction target and evaluation ground truth consistent with the original task.

\begin{figure}[!t]
\centering
\includegraphics[width=\linewidth]{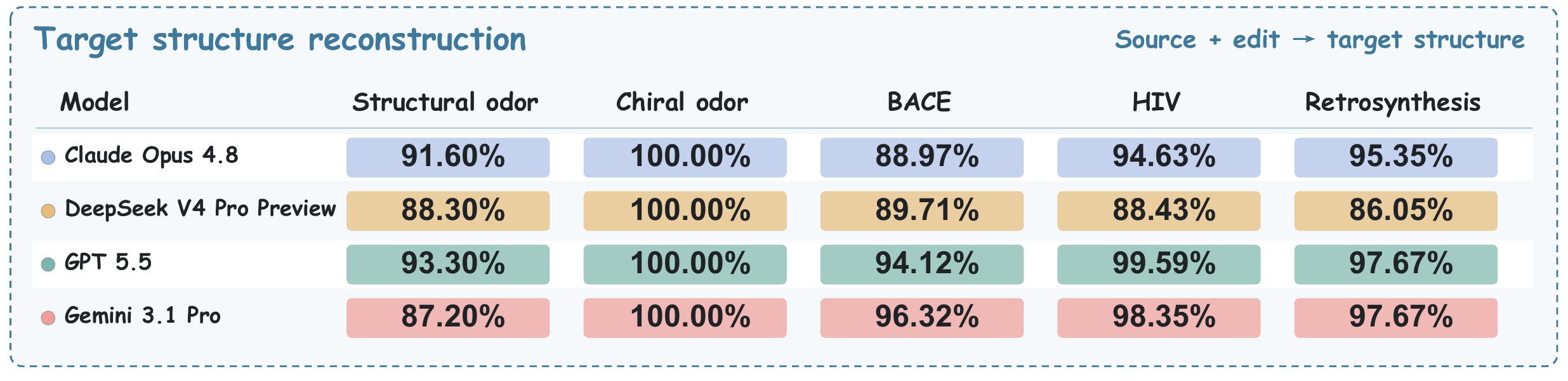}
\renewcommand{\thefigure}{2}
\caption{Validation of target structure reconstruction under the identity-blinded setting. The LLM reconstructs the complete SMILES of the target molecule from the source molecule structure and the structural change description. The reaction yield task involves multicomponent formulations, for which single-molecule structure reconstruction is not applicable.}
\label{fig:2}
\end{figure}

To verify whether the identity-blinded setting preserves sufficient information to define the target, we construct structure reconstruction tasks corresponding to the five molecular tasks. Structure reconstruction accuracy ranges from 86.05\% to 100\%, and all four models correctly recover every chiral target molecule, as shown in Figure~\ref{fig:2}. The reconstruction results are not provided in subsequent property prediction experiments, in which models predict directly from the source structure and edit information.

\subsection{Source Label Conditions and Prediction Reference}

As additional contextual information, source entity property labels can be used to analyze whether source entity properties influence model predictions. We control this information in the C1 condition through the C3 condition and use the C0 condition, which directly provides the target structure, as a reference, as shown in Figure~\ref{fig:1} and Table~\ref{tab:1}. Both the source structure and source label in the input are explicitly attributed to the source entity. The C2 condition and the C3 condition retain the same source structure, edit description, prediction target, and evaluation ground truth, differing only in the source label. False source labels are constructed according to the output type. Binary labels are inverted, alternative label sets are used for set prediction, and numerical predictions use replacement values that differ sufficiently from the original values. Detailed rules are provided in Appendix A.

\begingroup
\setlength{\tabcolsep}{3pt}
\renewcommand{\arraystretch}{1.0}
\renewcommand{\thetable}{1}
\begin{table}[t]
\caption{Prediction conditions.}
\label{tab:1}
\centering
\begin{tabular}{@{}c p{270pt}@{}}
\toprule
\textbf{Condition} & \textbf{Information provided to the model} \\
\midrule
C0 & Target structure representation (e.g., SMILES) \\
C1 & Source structure and edit description \\
C2 & Source structure, edit description, and true source label $y_s$ \\
C3 & Source structure, edit description, and false source label $\widetilde y_s$ \\
\bottomrule
\end{tabular}
\end{table}
\endgroup

\subsection{Label Group Analysis and Source Label Intervention}

During analysis, we divide samples into a label-consistent group and a label-inconsistent group according to the relationship between the true source and target labels. Binary classification compares whether labels are identical, whereas set-valued and numerical tasks use label overlap and property differences, respectively. Detailed rules are provided in Appendix A. If LLMs inappropriately use source molecule information for behaviors such as retrieval, the label-consistent group is likely to obtain higher scores. We therefore analyze model dependence on source molecule information. We report group scores and their ratio, the collapse ratio $\mathrm{CR}$, defined so that $\mathrm{CR}>1$ always indicates better performance in the label-consistent group, as shown in Figure~\ref{fig:1}\,\textup{(b)}.

We then compare the C2 condition and the C3 condition in target-matched pairs to measure whether model predictions change with source label replacement. For binary classification, we measure the proportion of predictions that switch to the replacement labels. For set prediction, we measure the change in preference between the two source label sets. For numerical prediction, we measure the proportion of predictions that change in the same direction as the source labels. Together, these three types of metrics characterize the influence of source label intervention on target predictions. Definitions are provided in Appendix B.

\section{Experiments}

\subsection{Experimental Setup}

The main diagnostic covers six tasks across four domains, including odor prediction, molecular property prediction, retrosynthesis, and reaction yield prediction, as listed in Table~\ref{tab:2}. The odor data follow the data curation procedure of \citep{su2026nose}, with structural change and chirality change tasks constructed from Pyrfume data.\footnote{\url{https://github.com/pyrfume/pyrfume-data}} The molecular property, retrosynthesis, and yield data are from \citep{wu2018moleculenet}, \citep{liu2017retrosynthetic}, and \citep{ahneman2018predicting}, respectively. Data sources and construction procedures are described in Appendix A.

\begingroup
\setlength{\tabcolsep}{3pt}
\renewcommand{\arraystretch}{1.0}
\renewcommand{\thetable}{2}
\begin{table}[t]
\caption{Tasks and sample sizes in the main diagnostic.}
\label{tab:2}
\centering
\begin{tabular}{
@{}
>{\raggedright\arraybackslash}p{65pt}
>{\raggedright\arraybackslash}p{75pt}
>{\centering\arraybackslash}p{20pt}
>{\centering\arraybackslash}p{35pt}
>{\raggedright\arraybackslash}p{110pt}
>{\centering\arraybackslash}p{55pt}
@{}
}
\toprule
\textbf{Task} & 
\textbf{Label} & 
\textbf{N} & 
\textbf{C/I} & 
\textbf{Criterion} & 
\textbf{Metric} \\
\midrule

Structural odor & Descriptor set & 179 & 122/57 & Label overlap & F1 \\

Chiral odor & Descriptor set & 56 & 24/32 & Label overlap & F1 \\

BACE & Binary label & 136 & 68/68 & Activity agreement & Accuracy \\

HIV & Binary label & 242 & 121/121 & Activity agreement & Accuracy \\

Retrosynthesis & Reactant SMILES & 42 & 28/14 & Fingerprint similarity $\geq$0.5 & Exact match \\

Reaction yield & Continuous value & 100 & 50/50 & Yield gap $\leq$15\% & MAE \\

\bottomrule
\end{tabular}
\end{table}
\endgroup

The main diagnostic compares Claude-\allowbreak{}Opus-\allowbreak{}4.8, DeepSeek-\allowbreak{}V4-\allowbreak{}Pro-\allowbreak{}Preview, GPT-\allowbreak{}5.5, and Gemini-\allowbreak{}3.1-\allowbreak{}Pro, with an independent call for each sample. Odor prediction uses the mean of per-sample descriptor F1 scores, binary classification uses accuracy, retrosynthesis uses the exact match rate of canonicalized reactant SMILES representations, and yield prediction uses mean absolute error, MAE, in percentage points. Model calling procedures are described in Appendix C, and complete results are provided in Appendix D. The contextual evidence augmentation experiment covers eight odor-related tasks and compares Gemini-\allowbreak{}3.1-\allowbreak{}Pro, DeepSeek-\allowbreak{}V4-\allowbreak{}Pro-\allowbreak{}Preview, Gemma-\allowbreak{}4-\allowbreak{}31B, Qwen3.5-\allowbreak{}27B \citep{qwen35blog}, and Qwen3.5-\allowbreak{}9B. Direct LLM prediction and external evidence augmentation use the same questions and evaluation answers. Tool-based contextual evidence provides odor or receptor matching information through NOSE, while RDKit implements structural and chirality change operations. Task sizes, data sources and processing, and tool functions are described in Appendix E.

\subsection{Benchmark Prediction under the Target-Identity-Known Condition}

The C0 condition provides a prediction reference with conventional target inputs, corresponding to the conventional benchmark prediction setting, as shown in Table~\ref{tab:3}. Model performance varies by task. BACE and HIV are binary classification tasks with a random baseline of approximately 50\%, and most models have accuracy near this baseline on these two tasks. Claude-\allowbreak{}Opus-\allowbreak{}4.8 also achieves accuracy slightly above the random baseline on BACE and HIV, as does Gemini-\allowbreak{}3.1-\allowbreak{}Pro on HIV. Under conventional benchmark evaluation, these results reflect model prediction performance across tasks. Retrosynthesis exact match rates range from 34.9\% to 48.8\%, indicating that models can generate predictions consistent with the true reactants for some samples. Yield MAE ranges from 32.34 to 56.46 percentage points, exceeding the MAE of 21.98 for the mean baseline and 21.65 for the median baseline. Model predictions therefore still do not outperform simple statistical baselines under the current setting. The subsequent diagnostic further examines source information dependence in predictions using the original evaluation answers. The diagnostic covers models with different levels of performance across tasks.

\begingroup
\setlength{\tabcolsep}{4pt}
\renewcommand{\arraystretch}{1.0}
\renewcommand{\thetable}{3}
\begin{table}[t]
\caption{Prediction results under the target-identity-known condition C0.}
\label{tab:3}
\centering
\begin{tabular}{
@{}
>{\raggedright\arraybackslash}p{120pt}
>{\centering\arraybackslash}p{55pt}
>{\centering\arraybackslash}p{65pt}
>{\centering\arraybackslash}p{55pt}
>{\centering\arraybackslash}p{65pt}
@{}
}
\toprule
\textbf{Task} & 
\textbf{Claude} & 
\textbf{DeepSeek} & 
\textbf{GPT} & 
\textbf{Gemini} \\
\midrule

Structural odor F1 
& 0.326 & 0.365 & 0.368 & 0.400 \\

Chiral odor F1 
& 0.198 & 0.250 & 0.234 & 0.272 \\

BACE accuracy 
& 57.40\% & 51.50\% & 51.50\% & 51.50\% \\

HIV accuracy 
& 57.90\% & 31.40\% & 34.30\% & 61.20\% \\

Retrosynthesis EM 
& 41.90\% & 34.90\% & 48.80\% & 44.20\% \\

Yield MAE 
& 32.34 & 56.46 & 48.00 & 35.04 \\

\bottomrule
\end{tabular}
\end{table}
\endgroup

\subsection{Source Molecular Information Dependence Analysis}

\begin{figure}[!htbp]
\centering
\includegraphics[width=\linewidth]{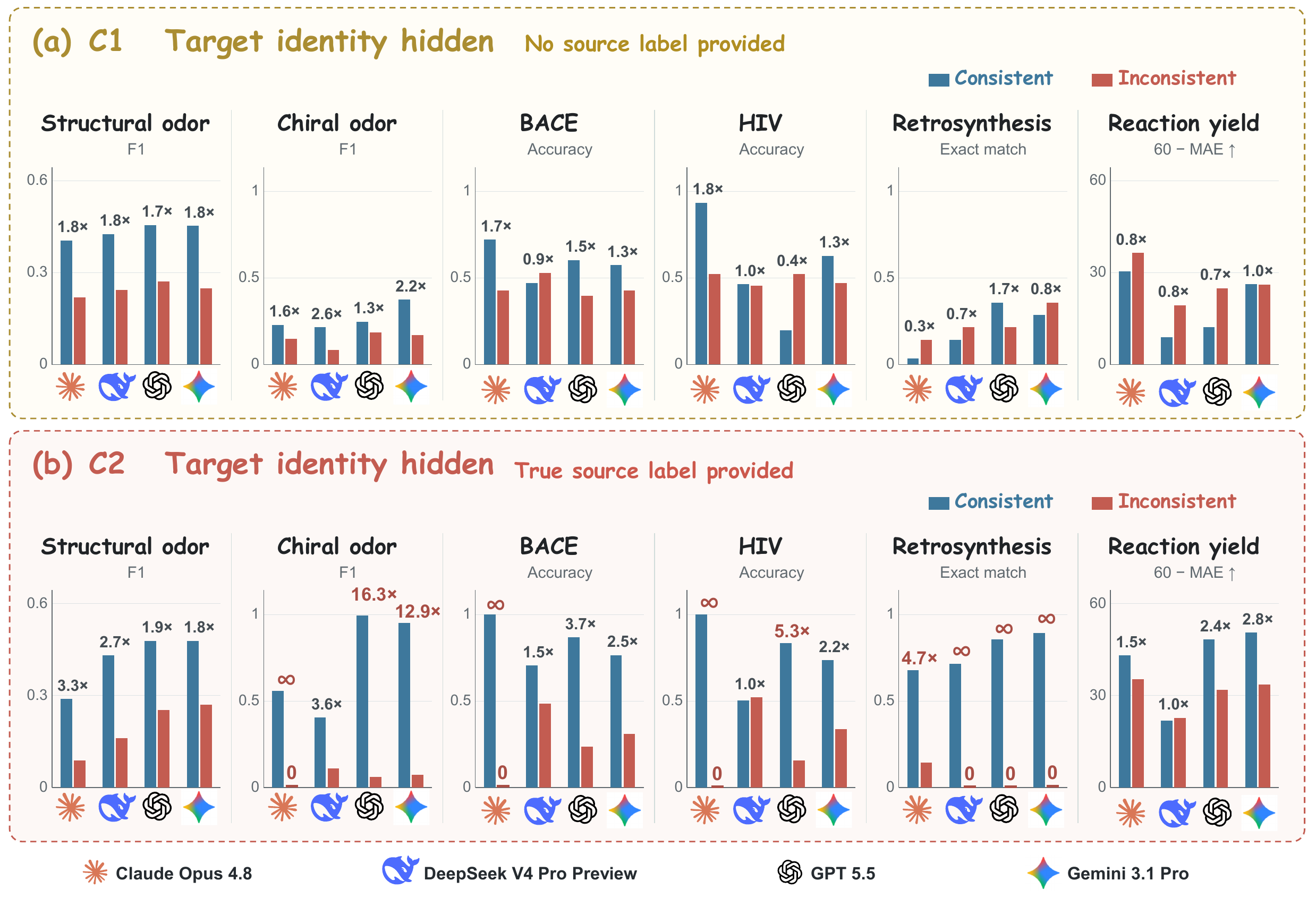}
\renewcommand{\thefigure}{3}
\caption{Source information dependence analysis. (a) C1 provides only source structures and edit descriptions, and compares label-consistent and label-inconsistent groups. (b) C2 additionally provides true source labels and evaluates the same group comparison. Numbers above bars indicate collapse ratios (CR).}
\label{fig:3}
\end{figure}

Under the C1 condition, which provides only the source molecule structure and edit description, all four models obtain higher F1 in the label-consistent group on both odor tasks, and three models show the same trend on BACE, as shown in Figure~\ref{fig:3}a. Overall, 10 of the 24 model--task combinations show a significant advantage for the label-consistent group, indicating that even without source labels, some model predictions are already influenced by source--target property consistency. However, nearly half of the tasks still show no evident source molecule information dependence, indicating that this pattern is not yet consistently observed across all tasks when only the structure and edit description are provided. Under the C2 condition, which additionally provides the true source molecule labels, the between-group differences increase substantially, as shown in Figure~\ref{fig:3}b. The label-consistent group has an advantage in 22 of the 24 model--task combinations, and performance in the label-inconsistent group approaches complete collapse on several tasks. Claude-\allowbreak{}Opus-\allowbreak{}4.8 achieves 100\% accuracy in the label-consistent group on BACE and HIV, whereas accuracy in the label-inconsistent group falls to 0\%. In the chirality change odor task, GPT-\allowbreak{}5.5 obtains F1 scores of 0.994 and 0.061 in the label-consistent and label-inconsistent groups, respectively, while Gemini-\allowbreak{}3.1-\allowbreak{}Pro obtains 0.952 and 0.074. On many tasks, the label-inconsistent group metric falls to 0, yielding an infinite CR. These results indicate that when source labels are provided as explicit context, model predictions are more readily influenced by source labels, obtaining apparently high performance when source and target labels agree but carry source information over to the target when properties change, leading to prediction errors. Statistical significance is reported in Appendix F.

\begin{figure}[!htbp]
\centering
\includegraphics[width=\linewidth]{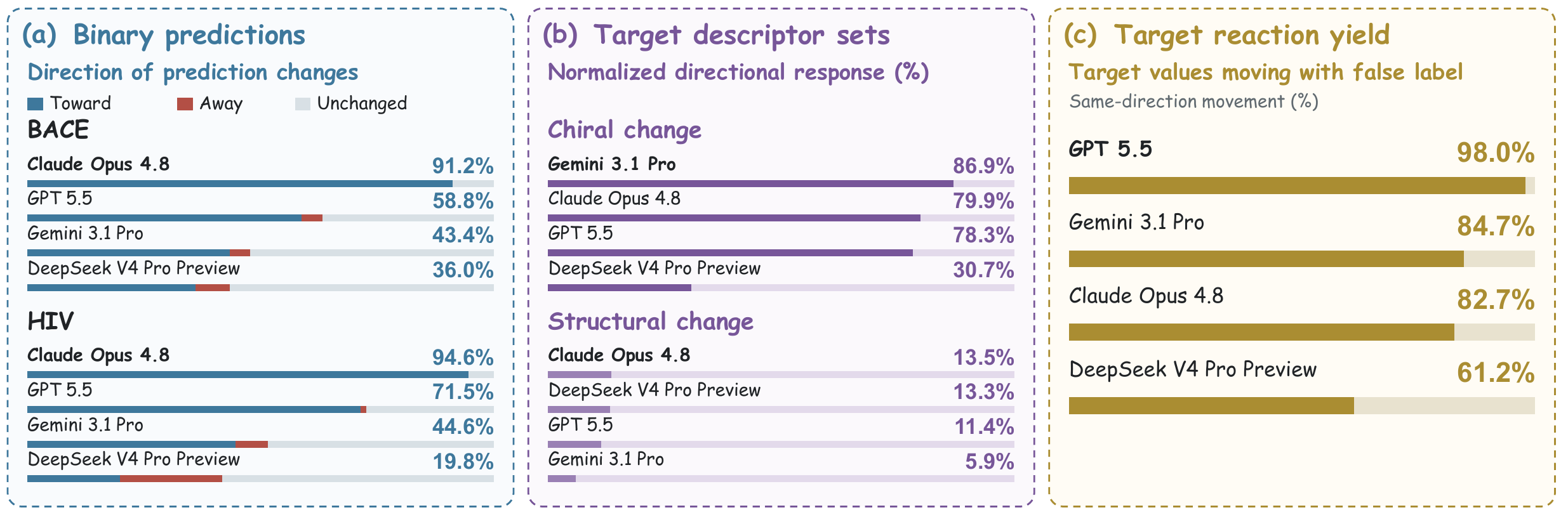}
\renewcommand{\thefigure}{4}
\caption{Effects of source label intervention on target prediction. (a) Binary prediction changes after replacing only the source label while keeping the target and other context fixed. Toward indicates movement toward the replacement source label, Away indicates movement away from it, and Unchanged indicates no prediction change. (b) Normalized directional response for set-valued predictions. (c) Same-direction response rate for numerical predictions, measuring whether prediction changes follow the direction of the replacement source label.}
\label{fig:4}
\end{figure}

\subsection{Source Molecular Label Interventions}

Starting from the C2 condition, we further hold the source molecule structure, edit description, prediction target, and evaluation ground truth fixed, replacing only the true source labels with false source labels in the C3 condition to directly measure the influence of source label intervention on target predictions. In the binary classification tasks, 91.2\% of Claude-\allowbreak{}Opus-\allowbreak{}4.8 predictions switch the replacement source labels, compared with 58.8\%, 43.4\%, and 36.0\% for GPT-\allowbreak{}5.5, Gemini-\allowbreak{}3.1-\allowbreak{}Pro, and DeepSeek-\allowbreak{}V4-\allowbreak{}Pro-\allowbreak{}Preview, respectively, as shown in Figure~\ref{fig:4}a. Here, toward indicates that predictions switches to the false source label, while away indicates that predictions move away from the false source label. In set-valued output tasks, model predictions also tend to move toward the replacement labels. The normalized directional response ranges from 0.307 to 0.869 for the chirality change odor task, but only from 0.059 to 0.135 for the structural change odor task, as shown in Figure~\ref{fig:4}b. This difference indicates substantial variation across tasks in the extent to which models respond to source label intervention. In some tasks, prediction sets are strongly influenced by source labels, whereas in others, models still retain a degree of structural change information. In the yield prediction task, 61.2\% to 98.0\% of predicted values change in the same direction as the replacement source yields, as shown in Figure~\ref{fig:4}c. Because neither the target molecules nor the true evaluation values change, predictions changing in the direction of the source labels indicate a systematic influence of contextual source label information on model outputs. Overall, across binary, set-valued, and continuous output spaces, changing only the source labels can cause systematic changes in predictions for the same target entity, indicating that predictions observed under standard benchmark conditions may be influenced by source label context and are not determined entirely by target structural changes. Statistical significance is reported in Appendix F.

\begin{figure}[!htbp]
\centering
\includegraphics[width=\linewidth]{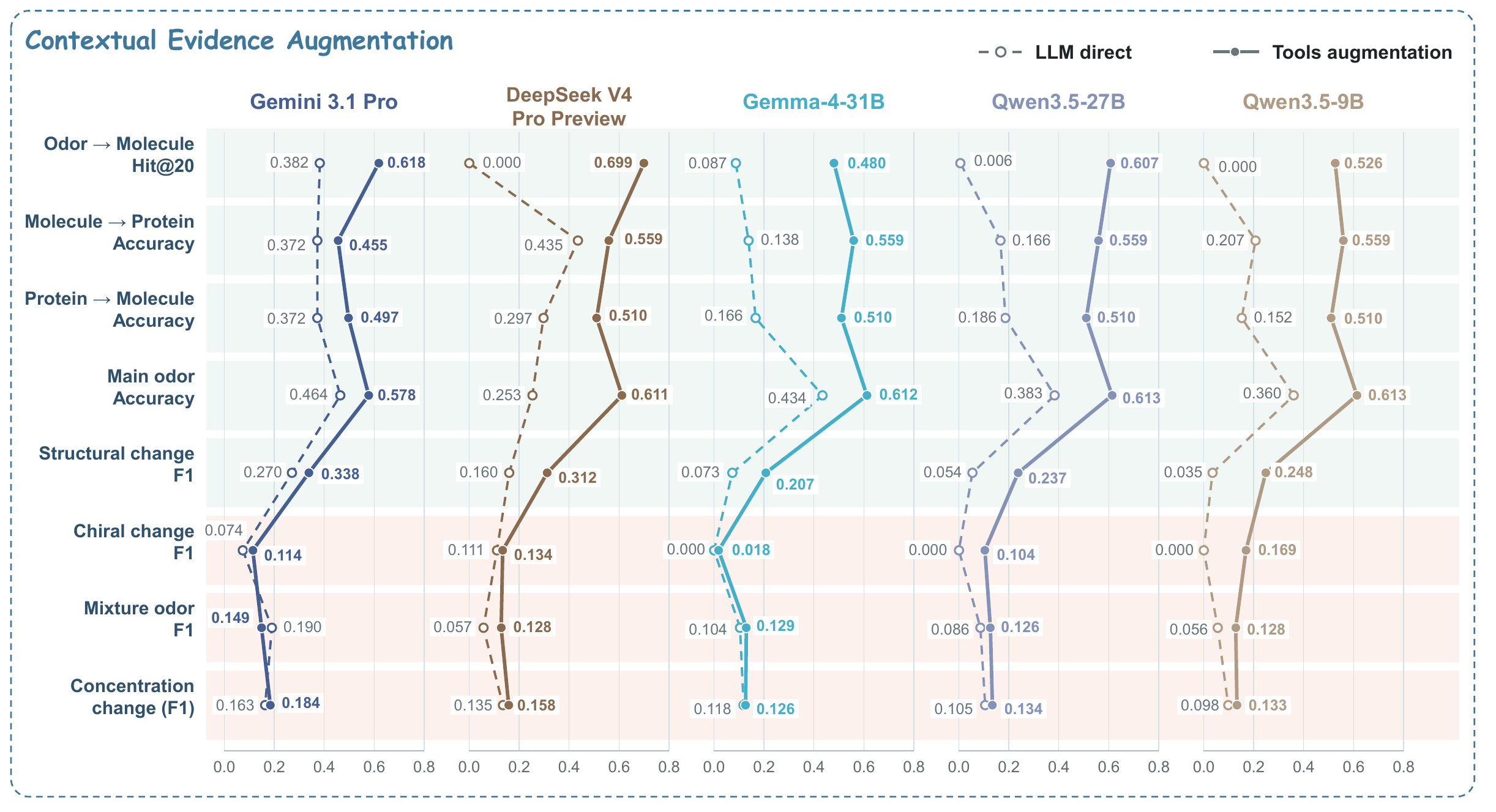}
\renewcommand{\thefigure}{5}
\caption{Effects of external predictive evidence augmentation on LLM prediction. Direct prediction is compared with prediction after adding task-relevant tool evidence. Columns denote models and rows denote odor-related tasks. Green and red regions indicate tasks with target-relevant evidence and limited evidence coverage, respectively.}
\label{fig:5}
\end{figure}

\subsection{Contextual Evidence Augmentation}

Source label intervention shows that contextual information can change models' judgments about targets. We further examine whether external evidence relevant to target prediction can guide models toward improved predictions and whether prediction gains are associated with evidence coverage.

The eight tasks are designed around the types of evidence the tools can provide. Odor retrieval, main odor selection, receptor matching, and structural change tasks have access to predictive evidence directed at the target. Chirality, mixture, and concentration tasks lack dedicated tool support and can obtain only approximate information from general odor tools. The chirality task uses odor predictions for structures after chirality inversion, the mixture task uses predictions from individual components or weighted representations, and the concentration task uses molecular odor predictions without concentration parameters. These three types of tasks involve enantiomer differences, component interactions, and concentration effects, which fall outside the dedicated modeling scope of the current tools. We compare LLM predictions with and without tool evidence, with results shown in Figure~\ref{fig:5} and other experimental settings described in Appendix E.

For tasks where targeted evidence is available, all five models improve to varying degrees. For example, on the structural change task, DeepSeek-\allowbreak{}V4-\allowbreak{}Pro-\allowbreak{}Preview improves from an F1 of 0.160 to 0.312, and Qwen3.5-\allowbreak{}9B from 0.035 to 0.248. Consistent improvements are also observed in odor retrieval, main odor selection, and receptor matching. In contrast, tasks with insufficient evidence coverage show limited gains or even declines. For example, Gemini-\allowbreak{}3.1-\allowbreak{}Pro's F1 on the mixture task falls from 0.190 to 0.149, indicating that information mismatched to the target property does not consistently improve predictions.

These results indicate that contextual information does not necessarily improve model performance by itself. Its effect depends on the relevance of the provided evidence to target prediction. Whereas source label intervention causes model outputs to change with labels, task-relevant external evidence can provide effective constraints and improve prediction on some tasks. These results further indicate that models are highly sensitive to contextual information and that reliable prediction requires evidence directly relevant to the target property.

\section{Conclusion}

This study examines whether high scores achieved by LLMs on chemical prediction benchmarks support a reliable structure--property reasoning capability. The C0 condition shows that LLM performance varies across tasks under conventional benchmark settings. The C1 condition and the C2 condition further show that, when the target structure can be determined from the source structure and edit description, some predictions are influenced by source-molecule-related information regardless of the model's benchmark performance, and this influence becomes stronger when source labels are explicitly provided. The paired intervention between the C2 condition and the C3 condition further shows that, on some tasks, changing only the source labels while holding the target entity fixed can cause systematic changes in predictions. These findings reveal the sensitivity of LLM chemical prediction results to contextual information and indicate that high benchmark scores alone are insufficient to show that models possess a reliable structure--property reasoning capability.

The identity-blinded diagnostic does not directly provide LLMs with target molecule identity information, but the source structure and edit description still preserve the information needed to define the target. This design applies not only to molecular property prediction. The reaction yield experiment also extends it to formulation systems comprising multiple components. For scientific prediction tasks in which targets can be defined by source entities and their edit operations, this framework can further analyze whether models can update predictions as entities change and whether reference information influences predictions.

The contextual evidence experiment further shows that, in tasks involving chemical prediction, task-relevant evidence from external tools can improve some predictions, with gains associated with the extent to which the evidence covers the target properties. This indicates that LLMs are highly sensitive to contextual information and that effective external evidence can provide additional constraints on prediction.

\FloatBarrier
\label{main-text-end}
\subsection*{AI use statement}

We used Cursor for coding assistance and Codex for language polishing and figure preparation. Gemini-3.1-Pro assisted with structural edit descriptions and concentration-label curation. Mixture odor labels were generated through two rounds of model calls and manually checked. The underlying structural edits were deterministically verified. The authors reviewed AI-assisted outputs and take responsibility for the final work.
\bibliography{references}
\bibliographystyle{iclr2027_conference}
\clearpage
\appendix
\FloatBarrier

\section{Diagnostic Data and Sample Construction}\label{app:A}

\begin{figure}[!htbp]
\centering
\includegraphics[width=\linewidth,height=0.67\textheight,keepaspectratio]{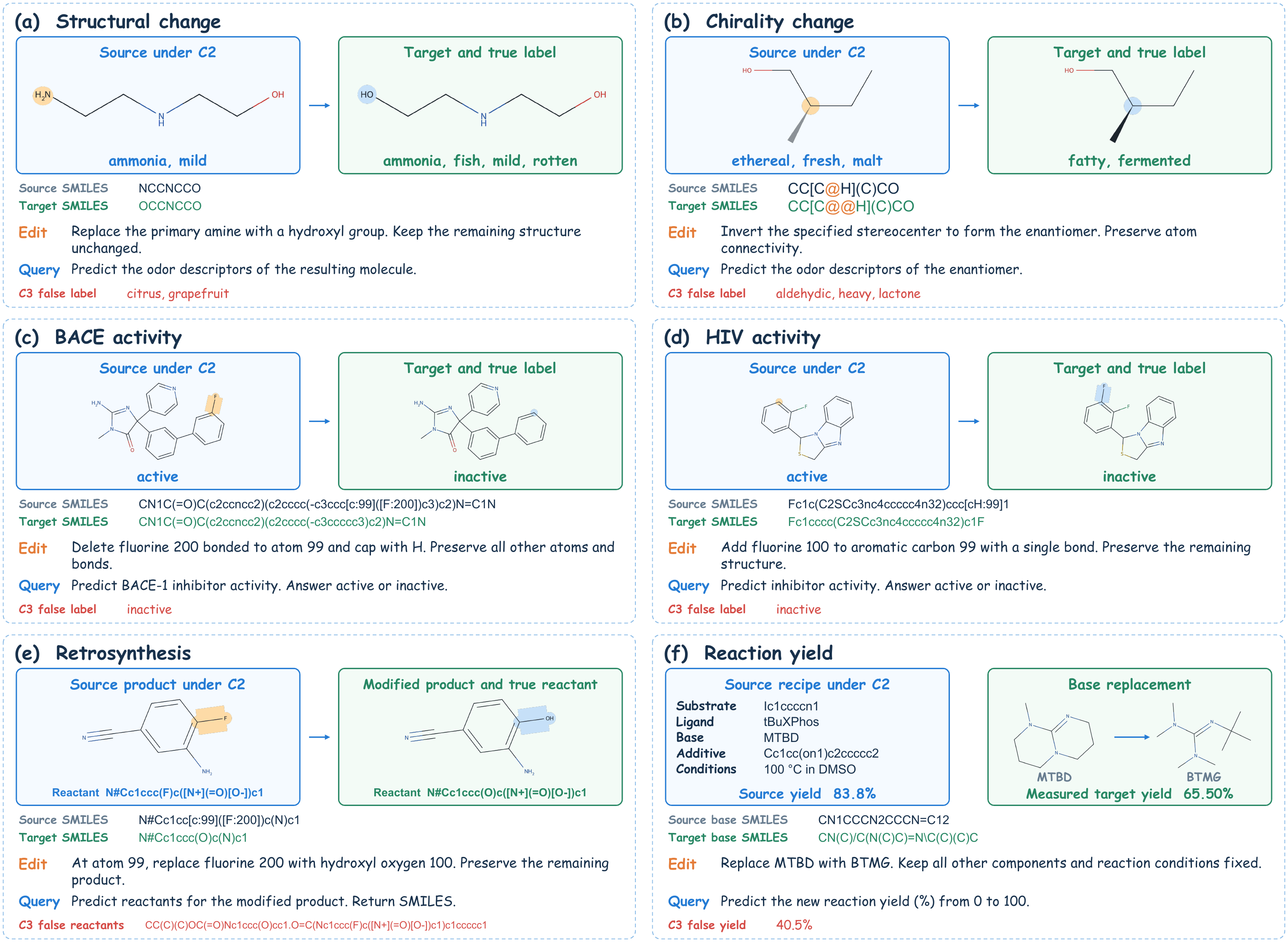}
\renewcommand{\thefigure}{A1}
\caption{Representative examples of the six diagnostic tasks. Each example shows the source molecule or reaction recipe, the natural-language edit instruction (Edit), the prediction query (Query), and the target structure or label after the edit. (a) Structural change odor prediction, where a terminal amine is replaced by a hydroxyl group. (b) Chirality change odor prediction, where a stereocenter is inverted while connectivity is preserved; red @ and @@ indicate chirality changes. (c) BACE activity prediction, where a fluorine substituent is removed and replaced by hydrogen. (d) HIV activity prediction, where a fluorine atom is added to an aromatic carbon. (e) Retrosynthesis prediction, where a fluorine substituent is replaced by a hydroxyl group and the corresponding reactants are predicted. (f) Reaction yield prediction, where MTBD is replaced by BTMG while other reaction conditions remain fixed. Orange and blue highlights indicate edited positions, and arrows indicate edit operations. The figure shows true source labels under C2 and false source labels under C3, while C1 provides no source label. All conditions share the same source entity, edit description, and prediction target. Target structures, Target SMILES, and target labels are shown only for illustration and evaluation and are not provided as model inputs. For yield prediction, the replaced component is part of the input condition, and the hidden target is the reaction yield.}
\label{fig:A1}
\end{figure}

\subsection{Olfactory Data}\label{app:A.1}

The olfactory data follow the data curation procedure of NOSE and are primarily drawn from the Mayhew, IFRA, Sigma-Aldrich, Flavornet, Arctander, GoodScents, and Leffingwell datasets included in Pyrfume. We used RDKit to standardize molecular structures, normalized the capitalization and word forms of descriptors, expanded compound labels, and deduplicated structure--descriptor pairs. The merged data contained 38,111 structure--label records covering 7,757 distinct structures. Following the existing held-out split and removing multicomponent structures yielded 6,985 single-component structures for diagnostic sample construction. For each structure, its odor labels were defined as the union of the retained descriptors associated with that structure.

\subsection{Structural and Chirality Changes}\label{app:A.2}

For the structural change task, we selected molecules with 4--60 heavy atoms and computed molecular similarity using 2048-bit Morgan fingerprints with a radius of 2. Candidate molecular pairs were retained if their Tanimoto similarity was at least 0.60 and their heavy atom counts differed by no more than 6. We used the maximum common substructure to locate the region of structural change and verified through a deterministic editing procedure whether the edit operation could uniquely generate the target structure. From 7,498 candidate molecular pairs, we selected 179 directed edit samples, covering changes such as functional group replacement, fragment deletion, and double-bond saturation. Gemini-\allowbreak{}3.1-\allowbreak{}Pro then generated natural-language edit descriptions from the source structures and verified edit operations. Based on the relationship between source and target odor labels, 122 samples were assigned to the label-consistent group and the remaining 57 to the label-inconsistent group. Representative examples are shown in Figure~\ref{fig:A1}\,\textup{(a)}. For the chirality change task, we selected strict enantiomer pairs from molecules with identical connectivity, requiring all tetrahedral stereocenters to be explicitly specified and molecular connectivity to remain unchanged after chirality inversion. We further filtered paired samples by the Jaccard similarity of their odor labels, retaining pairs with a Jaccard similarity of 0 or at least 0.8. This yielded 28 pairs and 56 bidirectional diagnostic samples, including 24 in the label-consistent group and 32 in the label-inconsistent group. The contextual evidence augmentation experiment used the same selection criteria. Examples of edit descriptions are shown in Figure~\ref{fig:A1}\,\textup{(b)}.

\subsection{Molecular Properties}\label{app:A.3}

The BACE and HIV tasks use binary activity labels from MoleculeNet. The original datasets contained 1,513 and 41,127 records, respectively. After checking for invalid, multicomponent, and duplicate structures, 1,513 and 38,034 single-component structures were retained. Source--target pairs were required to have a Morgan fingerprint Tanimoto similarity of at least 0.80 and a difference in heavy atom count of no more than 3. Edit operations were limited to fragment addition, deletion, or replacement at a single attachment site, with the changed fragment on either side containing at most 4 atoms. The target structure was verified after executing the edit, and the most similar source molecule was retained for each target. We then retained all label-inconsistent candidates and randomly sampled an equal number of label-consistent candidates, yielding 136 diagnostic samples for BACE and 242 for HIV. The edit descriptions used atom mapping to mark the specific operation sites while keeping the remaining structure and stereochemical information unchanged. Examples of edit descriptions are shown in Figure~\ref{fig:A1}\,\textup{(c)} and Figure~\ref{fig:A1}\,\textup{(d)}.

\subsection{Retrosynthesis}\label{app:A.4}

The retrosynthesis task uses the USPTO-50K test set curated by RetroXpert \citep{yan2020retroxpert}, which contains 5,007 reaction records. After removing the original atom mapping information, we standardized product structures and retrieved candidate source--target pairs based on Morgan fingerprint similarity, with Tanimoto similarity ranging from 0.40 to 0.90. We then verified fragment edit operations at a single attachment site, obtaining 43 executable samples. Under the diagnostic conditions, we excluded 1 sample whose source reactants directly contained the target product, leaving 42 diagnostic samples. For consistency grouping, samples were divided according to the Morgan fingerprint Tanimoto similarity between the source and target reactant sets. The 28 samples with a similarity of at least 0.5 were assigned to the label-consistent group, and the remaining 14 to the label-inconsistent group. The prediction target labels were the reactant sets in the original target reaction records. Examples of edit descriptions are shown in Figure~\ref{fig:A1}\,\textup{(e)}.

\subsection{Reaction Yield}\label{app:A.5}

The yield task uses the Buchwald--Hartwig experimental data reported by Ahneman et al. and the dataset curated by the Doyle group. It contains 3,600 reaction records covering 15 substrates, 4 ligands, 3 bases, and 20 additives.\footnote{\url{https://github.com/doyle-lab-ucla/ochem-data}} We constructed source--target pairs by replacing a single component in the reaction formulation while holding all other reaction conditions fixed. From 4,700 candidate pairs, we sampled 50 pairs with a yield difference of no more than 15 percentage points and 50 pairs with a difference greater than 15 percentage points, yielding 100 diagnostic questions. Model inputs included the source reaction formulation, the component replacement operation, and the fixed reaction conditions. The prediction target was the experimental yield under the edited conditions. Examples of edit descriptions are shown in Figure~\ref{fig:A1}\,\textup{(f)}.

\subsection{False Source Label Construction and Sample Sizes across Conditions}\label{app:A.6}

The C3 condition keeps the source structure, edit description, prediction target, and target ground truth fixed, replacing only the source label provided in the C2 condition. The replacement label is still identified as the property or reaction outcome of the source entity, and the model is not informed that the label is false. In all conditions, the label-consistent and label-inconsistent groups are defined by the relationship between the true source and target labels and are not reassigned according to the replacement source labels.

\textbf{Olfactory set labels.} A descriptor set from another sample was selected as the replacement label, with the requirement that it have no overlap with either the current true source label or the target label. All 179 structural change olfaction samples and all 56 chirality change olfaction samples were retained for paired comparisons between the C2 and C3 conditions. Label replacement changes only the reference odor set received by the model, without changing the target structure or target odor labels.

\textbf{BACE and HIV binary labels.} The true source label was inverted. For samples with inconsistent true source and target labels, the inverted source label equals the target label, which follows directly from the binary label space.

\textbf{Retrosynthesis reactant labels.} True reactant sets from other reactions in the same USPTO-50K test set were selected as replacement source labels. Candidate labels were required to have a Morgan fingerprint Tanimoto similarity of no more than 0.3 to both the true source reactant set and the target reactant set and to belong to a different reaction class. Among candidates satisfying these constraints, preference was given to those with similarities close to the upper limit. After filtering, all 42 samples had replacement labels satisfying the constraints.

\textbf{Numerical yield labels.} Only the yield value associated with the source formulation was modified. Replacement values were searched over the range of 5\%--95\% in increments of 0.1 percentage points, with the requirement that they differ from both the true source yield and the target ground truth by at least 25 percentage points. Among candidates satisfying the constraints, we selected values with a smaller shift from the source value and farther from the range boundaries. Because some samples had no replacement value satisfying the constraints, the C3 condition retained 98 samples, including 50 in the label-consistent group and 48 in the label-inconsistent group. Paired metrics and overall scores for the C2 and C3 conditions were computed on these 98 shared samples.

\begingroup
\setlength{\tabcolsep}{2pt}
\renewcommand{\arraystretch}{1.12}
\renewcommand{\thetable}{A3}
\begin{table}[H]
\caption{Sample sizes for each condition and paired analysis.}\label{tab:A3}
\centering
\begin{tabular}{@{}>{\raggedright\arraybackslash}p{117.020pt}>{\raggedright\arraybackslash}p{49.073pt}>{\raggedright\arraybackslash}p{67.947pt}>{\raggedright\arraybackslash}p{49.073pt}>{\raggedright\arraybackslash}p{49.073pt}>{\raggedright\arraybackslash}p{45.298pt}@{}}
\toprule
\textbf{Task} & \textbf{C0} & \textbf{Structure reconstruction} & \textbf{C1} & \textbf{C2} & \textbf{C3} \\
\midrule
Structural change olfaction & 162 & 179 & 179 & 179 & 179 \\
Chirality change olfaction & 56 & 56 & 56 & 56 & 56 \\
BACE & 136 & 136 & 136 & 136 & 136 \\
HIV & 242 & 242 & 242 & 242 & 242 \\
Retrosynthesis & 43 & 43 & 42 & 42 & 42 \\
Reaction yield & 100 & Not included & 100 & 100 & 98 \\
\bottomrule
\end{tabular}
\end{table}
\endgroup

\FloatBarrier

\section{Evaluation Metrics}\label{app:B}

\subsection{Task Performance and Between-Group Differences}\label{app:B.1}

The olfactory tasks use the mean of per-sample F1 scores. In the main diagnostic, descriptors are converted to lowercase and stripped of leading and trailing whitespace. For the predicted descriptor set $P_i$ and target descriptor set $Y_i$, the score is computed as

\[
F_{1,i}=\frac{2|P_i\cap Y_i|}{|P_i|+|Y_i|}
\]

The external evidence augmentation experiment additionally removes non-alphanumeric characters. Hits are computed from the number of matched labels, and precision and recall are computed using the lengths of the normalized label lists. Binary classification tasks use accuracy. The retrosynthesis task uses the exact match rate of standardized and sorted reactant sets. The yield task uses mean absolute error, denoted MAE. The odor retrieval task uses the proportion of queries for which the target appears among the top 20 results, denoted Hit@20.

For score-based metrics $S$, the between-group ratio is

\[
\mathrm{CR}=S_{\mathrm{consistent}}/S_{\mathrm{inconsistent}}
\]

For the error-based yield MAE metric, it is defined as

\[
\mathrm{CR}=\mathrm{MAE}_{\mathrm{inconsistent}}/\mathrm{MAE}_{\mathrm{consistent}}
\]

Under both definitions, CR > 1 indicates that the label-consistent group outperforms the label-inconsistent group. When the score of the label-inconsistent group is 0 and that of the label-consistent group is greater than 0, CR is infinite. The actual performance values of both groups are also reported, and an infinite CR is not used alone to interpret effect size.

\subsection{Directional Switching in Binary Classification}\label{app:B.2}

Let $p_{2,i}$ and $p_{3,i}$ denote the binary predictions for the same target sample under the C2 and C3 conditions, respectively, and let $\widetilde y_{s,i}$ denote the replacement source label provided in the C3 condition. The directional switching rate is defined as

\[
D_{\mathrm{binary}}=
\frac{1}{N}\sum_{i=1}^{N}
\mathbf{1}\{p_{2,i}\ne p_{3,i},\ p_{3,i}=\widetilde y_{s,i}\}
\]

The denominator $N$ is the number of all valid paired samples across the C2 and C3 conditions. This metric counts only samples for which the prediction changes from the C2 condition to the C3 condition and the C3 output switches to the replacement source label. If the C3 output matches the replacement source label but the prediction remains unchanged between the C2 and C3 conditions, it is not counted as a directional switch.

\subsection{Directional Response in Set Predictions}\label{app:B.3}

To measure whether the predicted set moves toward the replacement source label from the C2 condition to the C3 condition, we use Jaccard similarity \citep{jaccard1901etude} to measure the overlap between descriptor sets.

Let $A_i$ and $\widetilde A_i$ denote the true source descriptor set and the replacement source descriptor set provided in the C3 condition for sample $i$, respectively. Let $P_{2,i}$ and $P_{3,i}$ denote the descriptor sets predicted by the model for the same target molecule under the C2 and C3 conditions, respectively. Jaccard similarity is defined as

\[
J(U,V)=\frac{|U\cap V|}{|U\cup V|}
\]

The set directional response is defined as

\[
R_i=
\frac{
J(P_{3,i},A'_i)-J(P_{3,i},A_i)
-J(P_{2,i},A'_i)+J(P_{2,i},A_i)
}{2[1-J(A_i,A'_i)]}
\]

Here, $R_i>0$ indicates that the predicted set moves toward the replacement source label relative to the true source label from the C2 condition to the C3 condition. A value of $R_i<0$ indicates that the prediction moves in the opposite direction. A value of $R_i=0$ indicates that the relative preference of the two predictions for the two source label sets remains unchanged. When the prediction changes from exactly matching the true source set to exactly matching the replacement source set, $R_i=1$. In both olfactory tasks, the true source and replacement source sets are disjoint, so the denominator is always 2 for every sample. Figure~\ref{fig:4}b reports the mean of $R_i$ across all samples.

\subsection{Direction of Change in Numerical Predictions}\label{app:B.4}

For each valid paired sample $i$, let $y_{s,i}$ denote the true source formulation yield provided in the C2 condition and $\widetilde y_{s,i}$ the replacement source formulation yield in the C3 condition. Let $p_{2,i}$ and $p_{3,i}$ denote the yields predicted by the model for the same target formulation under the C2 and C3 conditions, respectively. With $N$ denoting the number of valid paired samples, the yield directional response rate is defined as

\[
D_{\mathrm{yield}} = \frac{1}{N} \sum_{i=1}^{N} \mathds{1} \left[ (p_{3,i}-p_{2,i}) (\widetilde y_{s,i}-y_{s,i})>0 \right]
\]

Here, $\mathds{1}[\cdot]$ is the indicator function, which equals 1 when the condition holds and 0 otherwise. The term $p_{3,i}-p_{2,i}$ represents the change in the predicted target yield between the C2 and C3 conditions, and $\widetilde y_{s,i}-y_{s,i}$ represents the change in the source yield label provided to the model. A positive product indicates that the direction of the prediction change matches the direction of the source label change. Samples with unchanged predictions remain in the denominator but are not counted as directional responses. This study has 98 valid paired samples across the C2 and C3 conditions, all with nonzero source label changes and parseable numerical predictions. All yields are expressed on a percentage scale, and differences are expressed in percentage points. Results are reported as $100D_{\mathrm{yield}}\%$. This metric measures only the direction of prediction changes and does not reflect their magnitude.

\FloatBarrier

\section{Prompt Settings and Experimental Protocol}\label{app:C}

All samples are independently submitted to the models. The C1 condition provides the source structure and edit description. The C2 condition additionally provides a true label explicitly identified as belonging to the source entity. The C3 condition replaces this label with a false source label. Depending on the task, models output descriptor sets, class labels, reactant SMILES, or numerical yields. Because Claude imposes safety restrictions on the HIV task, the prompts use a generic description of enzyme inhibitory activity in place of the HIV task name, while evaluation still uses the original HIV labels. The contextual evidence experiments compare direct prediction without tools with prediction augmented by tools. Tool results are provided to the model as contextual information, and the model may use these results to make further tool calls or produce a final answer. Structured messages are used for communication between models and tools. The direct prediction and tool-augmented conditions use the same question inputs and evaluation ground truth. Odor edit descriptions and concentration labels were curated with assistance from Gemini. Mixture labels were generated through two rounds of model calls and were manually extracted and checked. During evaluation, repeated call results were merged by sample identifier. Empty answers were counted as incorrect, and empty descriptor outputs received an F1 score of 0. When tool calls reached the interaction limit, the top 10 descriptors from the last valid odor prediction were used as the final output for some samples.

\FloatBarrier

\section{Complete Experimental Results}\label{app:D}

This section summarizes the complete results of the identity-blinded diagnostic. Table~\ref{tab:D1} reports the results of the independent structure reconstruction task, which verifies whether the source structure and edit description can effectively define the target structure. Table~\ref{tab:D2} summarizes paired response metrics under source label intervention between the C2 and C3 conditions, including binary classification switching, set directional response, and yield directional response, to analyze whether predictions change systematically with source label replacement. Table~\ref{tab:D3} further reports the absolute errors of the yield task under the C2 and C3 conditions on the same 98 paired samples, supplementing the analysis of the relationship between directional response and changes in overall error. Table~\ref{tab:D4} presents the complete performance of the label-consistent and label-inconsistent groups for each task under the C1 and C2 conditions. All groups are defined by the relationship between the true source and target labels.

\begingroup
\setlength{\tabcolsep}{4pt}
\renewcommand{\arraystretch}{1.12}
\renewcommand{\thetable}{D1}
\begin{table}[H]
\caption{Accuracy on the independent structure reconstruction task.}\label{tab:D1}
\centering
\begin{tabular}{@{}>{\raggedright\arraybackslash}p{72pt}>{\raggedright\arraybackslash}p{69.442pt}>{\raggedright\arraybackslash}p{80.407pt}>{\raggedright\arraybackslash}p{70pt}>{\raggedright\arraybackslash}p{69.442pt}@{}}
\toprule
\textbf{Task} & \textbf{Claude} & \textbf{DeepSeek} & \textbf{GPT} & \textbf{Gemini} \\
\midrule
Structural change & 164/179, 91.62\% & 158/179, 88.27\% & 167/179, 93.30\% & 156/179, 87.15\% \\
Chirality change & 56/56, 100\% & 56/56, 100\% & 56/56, 100\% & 56/56, 100\% \\
BACE & 121/136, 88.97\% & 122/136, 89.71\% & 128/136, 94.12\% & 131/136, 96.32\% \\
HIV & 229/242, 94.63\% & 214/242, 88.43\% & 241/242, 99.59\% & 238/242, 98.35\% \\
Retrosynthesis & 41/43, 95.35\% & 37/43, 86.05\% & 42/43, 97.67\% & 42/43, 97.67\% \\
\bottomrule
\end{tabular}
\end{table}
\endgroup

\begingroup
\setlength{\tabcolsep}{2pt}
\renewcommand{\arraystretch}{1.12}
\renewcommand{\thetable}{D2}
\begin{table}[H]
\caption{Paired response metrics under source label intervention between the C2 and C3 conditions.}\label{tab:D2}
\centering
\begin{tabular}{@{}>{\raggedright\arraybackslash}p{45pt}>{\raggedright\arraybackslash}p{45pt}>{\raggedright\arraybackslash}p{45pt}>{\raggedright\arraybackslash}p{80pt}>{\raggedright\arraybackslash}p{75pt}>{\raggedright\arraybackslash}p{45pt}@{}}
\toprule
\textbf{Model} & \textbf{BACE} & \textbf{HIV} & \textbf{Structural change} & \textbf{Chirality change} & \textbf{Yield} \\
\midrule
Claude & 91.2\% & 94.6\% & 0.135 & 0.799 & 82.7\% \\
DeepSeek & 36.0\% & 19.8\% & 0.133 & 0.307 & 61.2\% \\
GPT & 58.8\% & 71.5\% & 0.114 & 0.783 & 98.0\% \\
Gemini & 43.4\% & 44.6\% & 0.059 & 0.869 & 84.7\% \\
\bottomrule
\end{tabular}
\end{table}
\endgroup

\begingroup
\setlength{\tabcolsep}{4pt}
\renewcommand{\arraystretch}{1.12}
\renewcommand{\thetable}{D3}
\begin{table}[H]
\caption{Yield MAE on the same 98 paired samples.}\label{tab:D3}
\centering
\begin{tabular}{@{}>{\raggedright\arraybackslash}p{110pt}>{\raggedright\arraybackslash}p{45pt}>{\raggedright\arraybackslash}p{45pt}@{}}
\toprule
\textbf{Model} & \textbf{C2} & \textbf{C3} \\
\midrule
Claude-\allowbreak{}Opus-\allowbreak{}4.8 & 20.52 & 27.65 \\
DeepSeek-\allowbreak{}V4-\allowbreak{}Pro-\allowbreak{}Preview & 37.60 & 36.91 \\
GPT-\allowbreak{}5.5 & 19.43 & 33.15 \\
Gemini-\allowbreak{}3.1-\allowbreak{}Pro & 17.51 & 29.09 \\
\bottomrule
\end{tabular}
\end{table}
\endgroup

\begingroup
\setlength{\tabcolsep}{4pt}
\renewcommand{\arraystretch}{1.12}
\renewcommand{\thetable}{D4}
\begin{longtable}{@{}>{\raggedright\arraybackslash}p{89.636pt}>{\raggedright\arraybackslash}p{119.515pt}>{\raggedright\arraybackslash}p{82.167pt}>{\raggedright\arraybackslash}p{82.167pt}@{}}
\caption{Performance of the label-consistent/label-inconsistent groups under the C1 and C2 conditions.}\label{tab:D4}\\
\toprule
\textbf{Task} & \textbf{Model} & \textbf{C1} & \textbf{C2} \\
\midrule
\endfirsthead
\multicolumn{4}{l}{Table D4 continued}\\
\toprule
\textbf{Task} & \textbf{Model} & \textbf{C1} & \textbf{C2} \\
\midrule
\endhead
\midrule
\multicolumn{4}{r}{Continued on next page}\\
\endfoot
\bottomrule
\endlastfoot
Structural change & Claude-\allowbreak{}Opus-\allowbreak{}4.8 & 0.405 /\allowbreak{} 0.219 & 0.289 /\allowbreak{} 0.088 \\
Structural change & DeepSeek-\allowbreak{}V4-\allowbreak{}Pro-\allowbreak{}Preview & 0.426 /\allowbreak{} 0.243 & 0.431 /\allowbreak{} 0.160 \\
Structural change & GPT-\allowbreak{}5.5 & 0.455 /\allowbreak{} 0.271 & 0.479 /\allowbreak{} 0.254 \\
Structural change & Gemini-\allowbreak{}3.1-\allowbreak{}Pro & 0.453 /\allowbreak{} 0.248 & 0.478 /\allowbreak{} 0.270 \\
Chirality change & Claude-\allowbreak{}Opus-\allowbreak{}4.8 & 0.229 /\allowbreak{} 0.147 & 0.559 /\allowbreak{} 0.000 \\
Chirality change & DeepSeek-\allowbreak{}V4-\allowbreak{}Pro-\allowbreak{}Preview & 0.215 /\allowbreak{} 0.084 & 0.405 /\allowbreak{} 0.111 \\
Chirality change & GPT-\allowbreak{}5.5 & 0.247 /\allowbreak{} 0.186 & 0.994 /\allowbreak{} 0.061 \\
Chirality change & Gemini-\allowbreak{}3.1-\allowbreak{}Pro & 0.374 /\allowbreak{} 0.170 & 0.952 /\allowbreak{} 0.074 \\
BACE & Claude-\allowbreak{}Opus-\allowbreak{}4.8 & 0.721 /\allowbreak{} 0.426 & 1.000 /\allowbreak{} 0.000 \\
BACE & DeepSeek-\allowbreak{}V4-\allowbreak{}Pro-\allowbreak{}Preview & 0.471 /\allowbreak{} 0.529 & 0.706 /\allowbreak{} 0.485 \\
BACE & GPT-\allowbreak{}5.5 & 0.603 /\allowbreak{} 0.397 & 0.868 /\allowbreak{} 0.235 \\
BACE & Gemini-\allowbreak{}3.1-\allowbreak{}Pro & 0.574 /\allowbreak{} 0.426 & 0.765 /\allowbreak{} 0.309 \\
HIV & Claude-\allowbreak{}Opus-\allowbreak{}4.8 & 0.934 /\allowbreak{} 0.521 & 1.000 /\allowbreak{} 0.000 \\
HIV & DeepSeek-\allowbreak{}V4-\allowbreak{}Pro-\allowbreak{}Preview & 0.463 /\allowbreak{} 0.455 & 0.504 /\allowbreak{} 0.521 \\
HIV & GPT-\allowbreak{}5.5 & 0.198 /\allowbreak{} 0.521 & 0.835 /\allowbreak{} 0.157 \\
HIV & Gemini-\allowbreak{}3.1-\allowbreak{}Pro & 0.628 /\allowbreak{} 0.471 & 0.736 /\allowbreak{} 0.339 \\
Retrosynthesis & Claude-\allowbreak{}Opus-\allowbreak{}4.8 & 0.036 /\allowbreak{} 0.143 & 0.679 /\allowbreak{} 0.143 \\
Retrosynthesis & DeepSeek-\allowbreak{}V4-\allowbreak{}Pro-\allowbreak{}Preview & 0.143 /\allowbreak{} 0.214 & 0.714 /\allowbreak{} 0.000 \\
Retrosynthesis & GPT-\allowbreak{}5.5 & 0.357 /\allowbreak{} 0.214 & 0.857 /\allowbreak{} 0.000 \\
Retrosynthesis & Gemini-\allowbreak{}3.1-\allowbreak{}Pro & 0.286 /\allowbreak{} 0.357 & 0.893 /\allowbreak{} 0.000 \\
Yield & Claude-\allowbreak{}Opus-\allowbreak{}4.8 & 29.64 /\allowbreak{} 23.60 & 16.81 /\allowbreak{} 24.58 \\
Yield & DeepSeek-\allowbreak{}V4-\allowbreak{}Pro-\allowbreak{}Preview & 51.04 /\allowbreak{} 40.63 & 38.17 /\allowbreak{} 37.23 \\
Yield & GPT-\allowbreak{}5.5 & 47.83 /\allowbreak{} 35.12 & 11.68 /\allowbreak{} 28.03 \\
Yield & Gemini-\allowbreak{}3.1-\allowbreak{}Pro & 33.85 /\allowbreak{} 33.88 & 9.39 /\allowbreak{} 26.33 \\
\end{longtable}
\endgroup

\FloatBarrier

\section{Supplementary Experiments on Contextual Evidence}\label{app:E}

\subsection{Data and Evaluation Scope}\label{app:E.1}

Table~\ref{tab:E1} lists the task sizes, data sources, and evaluation metrics of the eight contextual evidence experiments. The structural change, chirality change, and concentration change tasks are all evaluated on subsets with disjoint source and target labels to avoid simple matching caused by label overlap.

\begingroup
\setlength{\tabcolsep}{4pt}
\renewcommand{\arraystretch}{1.12}
\renewcommand{\thetable}{E1}
\begin{table}[H]
\caption{Evaluation sets for the contextual evidence experiments.}\label{tab:E1}
\centering
\begin{tabular}{@{}>{\raggedright\arraybackslash}p{100.841pt}>{\raggedright\arraybackslash}p{134.455pt}>{\raggedright\arraybackslash}p{44.818pt}>{\raggedright\arraybackslash}p{93.371pt}@{}}
\toprule
\textbf{Task} & \textbf{Data source} & \textbf{Questions} & \textbf{Metric} \\
\midrule
Odor-to-mol retrieval & IFF & 173 & Hit@20 \\
Primary odor selection & IFRA 2019 & 1146 & Three-choice accuracy \\
Molecule-to-receptor & Emter et al. & 145 & Five-choice accuracy \\
Receptor-to-molecule & Emter et al. & 145 & Five-choice accuracy \\
Structural change olfaction & Molecular edit pairs from the main diagnostic & 57 & F1 \\
Chirality change olfaction & Enantiomer pairs from the main diagnostic & 32 & F1 \\
Mixture odor & Olfactorian & 319 & F1 \\
Concentration-dependent odor & Digitized Arctander materials & 362 & F1 \\
\bottomrule
\end{tabular}
\end{table}
\endgroup

\textbf{Primary odor.} We used the IFRA 2019 records curated by Pyrfume. Each record contains three odor descriptions associated with a molecule. The first-ranked descriptor is defined as the primary odor, and the other two serve as candidate options. The options were shuffled using random seed 42. This yielded 1,146 questions involving 1,060 distinct structures.

\textbf{Odor-to-molecule.} Odor descriptions were collected from the IFF ingredients compendium and linked to PubChem structural information through CAS numbers or chemical names. We retained 173 query samples with complete structures and descriptions and constructed a candidate pool of 147 distinct standardized structures. Models rank candidate molecules based on the odor descriptions.\footnote{\url{https://www.iff.com/scent/ingredients-compendium}}

\textbf{Bidirectional receptor matching.} We used the receptor library and activity data of Emter et al. \citep{emter2025decoding}, retaining their activity annotations. Ligand structures were retrieved from PubChem. Receptor sequences were translated from DNA sequences, and N-terminal tags were removed. After removing entries with missing structures or sequences and deduplicating the data, we obtained 179 valid pairs, including 145 active pairs involving 112 distinct molecules and 34 receptor sequences. Each active pair was used to construct five-choice tasks in both directions. Distractors were drawn from candidates not registered as active for the query, and option order was shuffled using random seed 42.

\textbf{Structural and chirality changes.} We used the source--target pairs and label definitions in Appendix A. Models receive the source structure, source odor labels, and edit requirements and are evaluated on 57 structural change questions and 32 chirality change questions with disjoint labels.

\textbf{Mixtures.} Component and proportion information was extracted from 503 public formulations on Olfactorian and linked to molecular structures through CAS numbers. We retained formulations containing at least two components for which parseable non-solvent materials accounted for at least 70\% of the formulation. Overall odor labels were extracted by a model from the names and descriptions, manually checked, and mapped to 753 standard descriptors. This yielded 319 formulations, of which 261 contained partially resolved components. Models receive component structures and relative proportions as input and predict the overall odor labels from the source text.\footnote{\url{https://olfactorian.com/}}

\textbf{Concentration changes.} We extracted records with explicit descriptions of odor differences between high and low concentrations from the digitized Arctander materials included in Pyrfume. Label cleaning yielded 212 records involving 210 distinct structures. Each record was used to construct questions in both directions, from high to low concentration and from low to high concentration, yielding 424 questions. Of these, 362 had disjoint labels on the two sides. Models receive the molecular structure and an odor description at one concentration and predict the odor change at the other concentration.

\subsection{Tool Evidence}\label{app:E.2}

NOSE provides matching scores between molecules and odors and between molecules and receptors. RDKit is used for structure standardization, functional group editing, and chirality inversion. Table~\ref{tab:E2} lists the tools and the information they provide. The contextual evidence experiments do not directly evaluate NOSE on its original tasks. Instead, NOSE is applied as an external tool in identity-blinded diagnostic settings, some of which extend beyond its original training and evaluation settings.

\begingroup
\setlength{\tabcolsep}{4pt}
\renewcommand{\arraystretch}{1.12}
\renewcommand{\thetable}{E2}
\begin{table}[H]
\caption{Information provided by the tools.}\label{tab:E2}
\centering
\begin{tabular}{@{}>{\raggedright\arraybackslash}p{83.927pt}>{\raggedright\arraybackslash}p{118.260pt}>{\raggedright\arraybackslash}p{179.298pt}@{}}
\toprule
\textbf{\textbf{Task}} & \textbf{\textbf{Tool evidence}} & \textbf{\textbf{Description}} \\
\midrule
\textbf{Primary odor and odor retrieval} & Molecule--descriptor matching scores and candidate rankings & NOSE training does not distinguish primary odors or assign different weights to the descriptors of a molecule. Task data are excluded from NOSE training to avoid data leakage. \\
\textbf{Bidirectional receptor matching} & Matching scores between query molecules and candidate receptors & The test set is independent of the NOSE training data to avoid data leakage. \\
\textbf{Structural changes} & Structures edited by RDKit and odor predictions from NOSE & The corresponding structural change test set is excluded from NOSE training to avoid data leakage. \\
\textbf{Chirality changes} & Structures inverted by RDKit and general odor predictions & NOSE models distance matrices and cannot distinguish chiral isomers. \\
\textbf{Mixtures} & Single-component odor predictions and retrieval using molecular representations weighted by component amounts & NOSE does not include a mixture input module. \\
\textbf{Concentration changes} & Molecular odor predictions without concentration parameters & NOSE does not include concentration-conditioned inputs. \\
\bottomrule
\end{tabular}
\end{table}
\endgroup

\Needspace{28\baselineskip}

\subsection{Tool Experiment Results}\label{app:E.3}

\begingroup
\setlength{\tabcolsep}{2pt}
\renewcommand{\arraystretch}{1.12}
\renewcommand{\thetable}{E3}
\begin{table}[H]
\caption{Prediction results for direct prediction and prediction augmented by tool evidence. Each cell reports the result of direct prediction without tools followed by the result after adding tool evidence. Task scope and sample sizes are given in Table~\ref{tab:E1}.}\label{tab:E3}
\centering
\begin{tabular}{@{}>{\raggedright\arraybackslash}p{80.047pt}>{\raggedright\arraybackslash}p{58.510pt}>{\raggedright\arraybackslash}p{58.835pt}>{\raggedright\arraybackslash}p{65.623pt}>{\raggedright\arraybackslash}p{58.623pt}>{\raggedright\arraybackslash}p{52.848pt}@{}}
\toprule
\textbf{Task} & \textbf{Gemini} & \textbf{DeepSeek} & \textbf{Gemma-\allowbreak{}4-\allowbreak{}31B} & \textbf{Qwen3.5-\allowbreak{}27B} & \textbf{Qwen3.5-\allowbreak{}9B} \\
\midrule
Odor-to-mol Hit@20 & 0.382 /\allowbreak{} 0.618 & 0.000 /\allowbreak{} 0.699 & 0.087 /\allowbreak{} 0.480 & 0.006 /\allowbreak{} 0.607 & 0.000 /\allowbreak{} 0.526 \\
Molecule-to-receptor accuracy & 0.372 /\allowbreak{} 0.455 & 0.435 /\allowbreak{} 0.559 & 0.138 /\allowbreak{} 0.559 & 0.166 /\allowbreak{} 0.559 & 0.207 /\allowbreak{} 0.559 \\
Receptor-to-molecule accuracy & 0.372 /\allowbreak{} 0.497 & 0.297 /\allowbreak{} 0.510 & 0.166 /\allowbreak{} 0.510 & 0.186 /\allowbreak{} 0.510 & 0.152 /\allowbreak{} 0.510 \\
Primary odor accuracy & 0.464 /\allowbreak{} 0.578 & 0.253 /\allowbreak{} 0.611 & 0.434 /\allowbreak{} 0.612 & 0.383 /\allowbreak{} 0.613 & 0.360 /\allowbreak{} 0.613 \\
Structural change F1 & 0.270 /\allowbreak{} 0.338 & 0.160 /\allowbreak{} 0.312 & 0.073 /\allowbreak{} 0.207 & 0.054 /\allowbreak{} 0.237 & 0.035 /\allowbreak{} 0.248 \\
Chirality change F1 & 0.074 /\allowbreak{} 0.114 & 0.111 /\allowbreak{} 0.134 & 0.000 /\allowbreak{} 0.018 & 0.000 /\allowbreak{} 0.104 & 0.000 /\allowbreak{} 0.169 \\
Mixture F1 & 0.190 /\allowbreak{} 0.149 & 0.057 /\allowbreak{} 0.128 & 0.104 /\allowbreak{} 0.129 & 0.086 /\allowbreak{} 0.126 & 0.056 /\allowbreak{} 0.128 \\
Concentration change F1 & 0.163 /\allowbreak{} 0.184 & 0.135 /\allowbreak{} 0.158 & 0.118 /\allowbreak{} 0.126 & 0.105 /\allowbreak{} 0.134 & 0.098 /\allowbreak{} 0.133 \\
\bottomrule
\end{tabular}
\end{table}
\endgroup

\FloatBarrier

\section{Statistical Significance}\label{app:F}

Between-group difference tests cover the C1 and C2 conditions. Each condition includes 6 tasks $\times$ 4 models, giving 24 model--task combinations and 48 comparisons across the two conditions. Each comparison tests the performance difference between the label-consistent and label-inconsistent groups. The BACE, HIV, and retrosynthesis tasks use classification or match-rate metrics. We apply two-sided Fisher's exact tests and report Cohen's h as the effect size. The structural change olfaction, chirality change olfaction, and reaction yield tasks use continuous or set-based performance metrics. We apply two-sided Mann--Whitney U tests and report Cliff's delta as the effect size. All 48 raw p-values undergo false discovery rate correction using the Benjamini--Hochberg method within a single comparison family. In the tables, q denotes the adjusted p-value, and the significance threshold is q < 0.05. For performance metrics, CR is defined as

\[
CR=S_{consistent}/S_{inconsistent}
\]

For error metrics, CR is defined as

\[
CR=E_{inconsistent}/E_{consistent}
\]

Both definitions use CR > 1 to indicate better performance in the label-consistent group. The 95\% confidence interval for CR is estimated from 10,000 within-group bootstrap resamples with replacement.

\subsection{Complete Tables of Between-Group Difference Tests}\label{app:F.1}

After Benjamini--Hochberg FDR correction, 34 of the 48 model--task comparisons reach significance at q < 0.05. Complete results are provided in Tables F1 and F2. In 6 comparisons under the C2 condition, the performance of the label-inconsistent group is 0. CR is undefined in these cases, but statistical tests and effect sizes are still reported. The C1 and C2 conditions use the same sample construction, with 42 paired samples for retrosynthesis and 100 samples for yield. Sample sizes for the other tasks are shown in the tables. The number of significant comparisons increases from 12/24 under the C1 condition to 22/24 under the C2 condition, indicating that performance differences between the label-consistent and label-inconsistent groups become more widespread when true source labels are provided.

\begingroup
\setlength{\tabcolsep}{2pt}
\renewcommand{\arraystretch}{1.12}
\renewcommand{\thetable}{F1}
\begin{longtable}{@{}>{\raggedright\arraybackslash}p{45pt}>{\raggedright\arraybackslash}p{38.152pt}>{\raggedright\arraybackslash}p{33.132pt}>{\raggedright\arraybackslash}p{33.132pt}>{\raggedright\arraybackslash}p{90.315pt}>{\raggedright\arraybackslash}p{56.224pt}>{\raggedright\arraybackslash}p{44.176pt}@{}}
\caption{Between-group difference tests under the C1 condition, with 12/24 significant comparisons. Subscripts c and i denote the label-consistent and label-inconsistent groups.}\label{tab:F1}\\
\toprule
\textbf{Model} & \textbf{$n_{\mathrm{c}}/n_{\mathrm{i}}$} & \textbf{$S_{\mathrm{c}}$} & \textbf{$S_{\mathrm{i}}$} & \textbf{CR and 95\% CI} & \textbf{$q$} & \textbf{Effect size} \\
\midrule
\endfirsthead
\multicolumn{7}{l}{Table F1 continued}\\
\toprule
\textbf{Model} & \textbf{$n_{\mathrm{c}}/n_{\mathrm{i}}$} & \textbf{$S_{\mathrm{c}}$} & \textbf{$S_{\mathrm{i}}$} & \textbf{CR and 95\% CI} & \textbf{$q$} & \textbf{Effect size} \\
\midrule
\endhead
\midrule
\multicolumn{7}{r}{Continued on next page}\\
\endfoot
\bottomrule
\endlastfoot
\addlinespace
\multicolumn{7}{@{}l}{\textbf{BACE}}\\*
Gemini & 68/68 & 57.4\% & 42.6\% & 1.34 [0.970, 1.950] & 0.1546 & 0.295 \\
GPT & 68/68 & 60.3\% & 39.7\% & 1.52 [1.065, 2.222] & 0.0361 & 0.415 \\
Claude & 68/68 & 72.1\% & 42.6\% & 1.69 [1.250, 2.400] & 0.0017 & 0.605 \\
DeepSeek & 68/68 & 47.1\% & 52.9\% & 0.89 [0.622, 1.250] & 0.6938 & -0.118 \\
\addlinespace
\multicolumn{7}{@{}l}{\textbf{HIV}}\\*
Gemini & 121/121 & 62.8\% & 47.1\% & 1.33 [1.063, 1.711] & 0.0307 & 0.317 \\
GPT & 121/121 & 19.8\% & 52.1\% & 0.38 [0.242, 0.550] & 7.68e-07 & -0.689 \\
Claude & 121/121 & 93.4\% & 52.1\% & 1.79 [1.520, 2.173] & 1.47e-12 & 1.009 \\
DeepSeek & 121/121 & 46.3\% & 45.5\% & 1.02 [0.772, 1.340] & 1.0000 & 0.017 \\
\addlinespace
\multicolumn{7}{@{}l}{\textbf{Chirality change olfaction}}\\*
Gemini & 24/32 & 37.4\% & 17.0\% & 2.20 [1.333, 3.826] & 0.0162 & 0.395 \\
GPT & 24/32 & 24.7\% & 18.6\% & 1.33 [0.837, 2.210] & 0.3044 & 0.176 \\
Claude & 24/32 & 22.9\% & 14.7\% & 1.56 [0.930, 2.696] & 0.0992 & 0.268 \\
DeepSeek & 24/32 & 21.5\% & 8.4\% & 2.56 [1.107, 7.558] & 0.0349 & 0.302 \\
\addlinespace
\multicolumn{7}{@{}l}{\textbf{Structural change olfaction}}\\*
Gemini & 122/57 & 45.3\% & 24.8\% & 1.83 [1.432, 2.463] & 1.55e-06 & 0.462 \\
GPT & 122/57 & 45.5\% & 27.1\% & 1.68 [1.363, 2.136] & 6.01e-06 & 0.434 \\
Claude & 122/57 & 40.5\% & 21.9\% & 1.85 [1.437, 2.508] & 1.07e-06 & 0.470 \\
DeepSeek & 122/57 & 42.6\% & 24.3\% & 1.75 [1.361, 2.385] & 5.22e-06 & 0.437 \\
\addlinespace
\multicolumn{7}{@{}l}{\textbf{Retrosynthesis}}\\*
Gemini & 28/14 & 28.6\% & 35.7\% & 0.80 [0.312, 2.500] & 0.7777 & -0.153 \\
GPT & 28/14 & 35.7\% & 21.4\% & 1.67 [0.625, 6.000] & 0.5681 & 0.319 \\
Claude & 28/14 & 3.6\% & 14.3\% & 0.25 [0.000, 1.000] & 0.3044 & -0.395 \\
DeepSeek & 28/14 & 14.3\% & 21.4\% & 0.67 [0.125, 3.000] & 0.7462 & -0.187 \\
\addlinespace
\multicolumn{7}{@{}l}{\textbf{Yield}}\\*
Gemini & 50/50 & 33.85 & 33.88 & 1.00 [0.761, 1.324] & 0.7777 & -0.042 \\
GPT & 50/50 & 47.83 & 35.12 & 0.73 [0.586, 0.913] & 0.0127 & 0.311 \\
Claude & 50/50 & 29.64 & 23.60 & 0.80 [0.632, 1.002] & 0.0992 & 0.206 \\
DeepSeek & 50/50 & 51.04 & 40.63 & 0.80 [0.628, 1.001] & 0.0972 & 0.210 \\
\end{longtable}
\endgroup

\begingroup
\setlength{\tabcolsep}{2pt}
\renewcommand{\arraystretch}{1.0}
\renewcommand{\thetable}{F2}
\begin{longtable}{@{}>{\raggedright\arraybackslash}p{45pt}>{\raggedright\arraybackslash}p{38.152pt}>{\raggedright\arraybackslash}p{33.132pt}>{\raggedright\arraybackslash}p{33.132pt}>{\raggedright\arraybackslash}p{90.315pt}>{\raggedright\arraybackslash}p{56.224pt}>{\raggedright\arraybackslash}p{44.176pt}@{}}
\caption{Between-group difference tests under the C2 condition, with 22/24 significant comparisons. Subscripts c and i denote the label-consistent and label-inconsistent groups.}
\label{tab:F2}\\
\toprule
\textbf{Model} & \textbf{$n_{\mathrm{c}}/n_{\mathrm{i}}$} & \textbf{$S_{\mathrm{c}}$} & \textbf{$S_{\mathrm{i}}$} & \textbf{CR [95\% CI]} & \textbf{$q$} & \textbf{Effect} \\
\midrule
\endfirsthead

\multicolumn{7}{l}{Table F2 continued}\\
\toprule
\textbf{Model} & \textbf{$n_{\mathrm{c}}/n_{\mathrm{i}}$} & \textbf{$S_{\mathrm{c}}$} & \textbf{$S_{\mathrm{i}}$} & \textbf{CR [95\% CI]} & \textbf{$q$} & \textbf{Effect} \\
\midrule
\endhead

\midrule
\multicolumn{7}{r}{Continued on next page}\\
\endfoot

\bottomrule
\endlastfoot

\addlinespace
\multicolumn{7}{@{}l}{\textbf{BACE}}\\*
Gemini & 68/68 & 76.5\% & 30.9\% & 2.48 [1.759, 3.846] & 4.91e-07 & 0.950 \\
GPT & 68/68 & 86.8\% & 23.5\% & 3.69 [2.500, 6.222] & 6.26e-13 & 1.384 \\
Claude & 68/68 & 100.0\% & 0.0\% & Undefined & 8.07e-39 & 3.142 \\
DeepSeek & 68/68 & 70.6\% & 48.5\% & 1.45 [1.103, 1.966] & 0.0226 & 0.454 \\

\addlinespace
\multicolumn{7}{@{}l}{\textbf{HIV}}\\*
Gemini & 121/121 & 73.6\% & 33.9\% & 2.17 [1.685, 2.933] & 4.40e-09 & 0.819 \\
GPT & 121/121 & 83.5\% & 15.7\% & 5.32 [3.689, 8.819] & 2.26e-26 & 1.489 \\
Claude & 121/121 & 100.0\% & 0.0\% & Undefined & 2.65e-70 & 3.142 \\
DeepSeek & 121/121 & 50.4\% & 52.1\% & 0.97 [0.754, 1.245] & 0.9168 & -0.033 \\

\addlinespace
\multicolumn{7}{@{}l}{\textbf{Chirality change olfaction}}\\*
Gemini & 24/32 & 95.2\% & 7.4\% & 12.94 [5.538, 152.82] & 1.43e-09 & 0.888 \\
GPT & 24/32 & 99.4\% & 6.1\% & 16.22 [8.110, 63.00] & 2.36e-11 & 1.000 \\
Claude & 24/32 & 55.9\% & 0.0\% & Undefined & 2.32e-06 & 0.583 \\
DeepSeek & 24/32 & 40.5\% & 11.1\% & 3.65 [1.579, 11.602] & 0.0361 & 0.287 \\

\addlinespace
\multicolumn{7}{@{}l}{\textbf{Structural change olfaction}}\\*
Gemini & 122/57 & 47.8\% & 27.0\% & 1.77 [1.386, 2.372] & 2.94e-06 & 0.449 \\
GPT & 122/57 & 47.9\% & 25.4\% & 1.89 [1.515, 2.465] & 4.69e-08 & 0.531 \\
Claude & 122/57 & 28.9\% & 8.8\% & 3.30 [2.125, 6.075] & 1.09e-06 & 0.446 \\
DeepSeek & 122/57 & 43.1\% & 16.0\% & 2.69 [1.963, 4.033] & 5.85e-11 & 0.630 \\

\addlinespace
\multicolumn{7}{@{}l}{\textbf{Retrosynthesis}}\\*
Gemini & 28/14 & 89.3\% & 0.0\% & Undefined & 5.15e-08 & 2.475 \\
GPT & 28/14 & 85.7\% & 0.0\% & Undefined & 2.14e-07 & 2.366 \\
Claude & 28/14 & 67.9\% & 14.3\% & 4.75 [1.800, 11.000] & 0.0046 & 1.161 \\
DeepSeek & 28/14 & 71.4\% & 0.0\% & Undefined & 1.36e-05 & 2.014 \\

\addlinespace
\multicolumn{7}{@{}l}{\textbf{Yield}}\\*
Gemini & 50/50 & 9.39 & 26.33 & 2.80 [1.833, 4.570] & 1.32e-08 & -0.690 \\
GPT & 50/50 & 11.68 & 28.03 & 2.40 [1.804, 3.298] & 4.50e-07 & -0.613 \\
Claude & 50/50 & 16.81 & 24.58 & 1.46 [1.098, 1.986] & 0.0100 & -0.322 \\
DeepSeek & 50/50 & 38.17 & 37.23 & 0.98 [0.735, 1.323] & 0.9035 & -0.020 \\

\end{longtable}
\endgroup

\subsection{Source Molecule Label Switching from C2 to C3}\label{app:F.2}

The C2 and C3 conditions use the same source entities, edit descriptions, prediction targets, and prompt templates. The only change is the replacement of the input source label from its true value with a counterfactual source label value. The three metrics follow the definitions in Sections B.2, B.3, and B.4 and cover binary, set-valued, and continuous output spaces. Results are shown in Table~\ref{tab:F3}. Binary classification tasks use the directional switching rate $D_{\mathrm{binary}}$, which measures the proportion of samples whose C3 output equals the replacement source label. Confidence intervals are computed using the Clopper--Pearson exact method. Set-valued tasks use the mean normalized Jaccard directional response RR. Positive values indicate that the predicted set moves toward the replacement source label relative to the true source label. Confidence intervals are estimated using bootstrap resampling. The numerical task uses the same-direction response rate $D_{\mathrm{yield}}$, which measures the proportion of samples for which the direction of the prediction change matches the direction of the source label change. Samples with unchanged predictions remain in the denominator but are not counted as same-direction responses. Confidence intervals are computed using the Clopper--Pearson exact method. The true source and replacement source descriptor sets are disjoint for all olfactory samples, so the normalization denominator of RR is always 2, allowing directional response values to be compared directly between the two olfactory tasks. In the chirality change olfaction task, 1 C3 output from Claude-\allowbreak{}Opus-\allowbreak{}4.8 could not be parsed. This sample was excluded from paired analysis, yielding a final n=55. All 98 paired samples in the yield task produced parseable predictions. The numbers of samples with unchanged predictions between the C2 and C3 conditions were 2 for Gemini-\allowbreak{}3.1-\allowbreak{}Pro, 0 for GPT-\allowbreak{}5.5, 2 for Claude-\allowbreak{}Opus-\allowbreak{}4.8, and 6 for DeepSeek-\allowbreak{}V4-\allowbreak{}Pro-\allowbreak{}Preview. Under the metric definition, these samples are not counted as same-direction responses. All confidence intervals for binary switching rates and set directional responses lie above the no-response level. For the yield task, all directional response confidence intervals lie above the random same-direction response reference value of 50\%.

\Needspace{10\baselineskip}
\begingroup
\setlength{\tabcolsep}{2pt}
\renewcommand{\arraystretch}{1.12}
\renewcommand{\thetable}{F3}
\begin{longtable}{@{}>{\raggedright\arraybackslash}p{80.947pt}>{\raggedright\arraybackslash}p{80.371pt}>{\raggedright\arraybackslash}p{49.073pt}>{\raggedright\arraybackslash}p{24.537pt}>{\raggedright\arraybackslash}p{30.298pt}>{\raggedright\arraybackslash}p{75.259pt}@{}}
\caption{Prediction response metrics and confidence intervals under source label intervention between the C2 and C3 conditions.}\label{tab:F3}\\
\toprule
\textbf{Task} & \textbf{Model} & \textbf{Metric} & \textbf{n} & \textbf{Value} & \textbf{95\% CI} \\
\midrule
\endfirsthead
\multicolumn{6}{l}{Table F3 continued}\\
\toprule
\textbf{Task} & \textbf{Model} & \textbf{Metric} & \textbf{n} & \textbf{Value} & \textbf{95\% CI} \\
\midrule
\endhead
\midrule
\multicolumn{6}{r}{Continued on next page}\\
\endfoot
\bottomrule
\endlastfoot
BACE & Gemini-\allowbreak{}3.1-\allowbreak{}Pro & D\_binary & 136 & 43.4\% & [34.9\%, 52.1\%] \\
BACE & GPT-\allowbreak{}5.5 & D\_binary & 136 & 58.8\% & [50.1\%, 67.2\%] \\
BACE & Claude-\allowbreak{}4.8-\allowbreak{}Opus & D\_binary & 136 & 91.2\% & [85.1\%, 95.4\%] \\
BACE & DeepSeek-\allowbreak{}V4-\allowbreak{}Pro & D\_binary & 136 & 36.0\% & [28.0\%, 44.7\%] \\
HIV & Gemini-\allowbreak{}3.1-\allowbreak{}Pro & D\_binary & 242 & 44.6\% & [38.3\%, 51.1\%] \\
HIV & GPT-\allowbreak{}5.5 & D\_binary & 242 & 71.5\% & [65.4\%, 77.1\%] \\
HIV & Claude-\allowbreak{}4.8-\allowbreak{}Opus & D\_binary & 242 & 94.6\% & [91.0\%, 97.1\%] \\
HIV & DeepSeek-\allowbreak{}V4-\allowbreak{}Pro & D\_binary & 242 & 19.8\% & [15.0\%, 25.4\%] \\
Structural change & Gemini-\allowbreak{}3.1-\allowbreak{}Pro & Mean $R$ & 179 & 0.059 & [0.039, 0.083] \\
Structural change & GPT-\allowbreak{}5.5 & Mean $R$ & 179 & 0.114 & [0.091, 0.138] \\
Structural change & Claude-\allowbreak{}4.8-\allowbreak{}Opus & Mean $R$ & 179 & 0.135 & [0.102, 0.170] \\
Structural change & DeepSeek-\allowbreak{}V4-\allowbreak{}Pro & Mean $R$ & 179 & 0.133 & [0.109, 0.157] \\
Chirality change & Gemini-\allowbreak{}3.1-\allowbreak{}Pro & Mean $R$ & 56 & 0.869 & [0.795, 0.934] \\
Chirality change & GPT-\allowbreak{}5.5 & Mean $R$ & 56 & 0.783 & [0.687, 0.874] \\
Chirality change & Claude-\allowbreak{}4.8-\allowbreak{}Opus & Mean $R$ & 55 & 0.799 & [0.726, 0.867] \\
Chirality change & DeepSeek-\allowbreak{}V4-\allowbreak{}Pro & Mean $R$ & 56 & 0.307 & [0.201, 0.416] \\
Yield & Gemini-\allowbreak{}3.1-\allowbreak{}Pro & D\_yield & 98 & 84.7\% & [76.0\%, 91.2\%] \\
Yield & GPT-\allowbreak{}5.5 & D\_yield & 98 & 98.0\% & [92.8\%, 99.8\%] \\
Yield & Claude-\allowbreak{}4.8-\allowbreak{}Opus & D\_yield & 98 & 82.7\% & [73.7\%, 89.6\%] \\
Yield & DeepSeek-\allowbreak{}V4-\allowbreak{}Pro & D\_yield & 98 & 61.2\% & [50.8\%, 70.9\%] \\
\end{longtable}
\endgroup

\end{document}

%% file: authors.tex
\author[1,2,\textdaggerdbl]{Yanyi Su}
\author[2,\textdaggerdbl]{Yu Zeng}
\author[2]{Hongshuai Wang}
\author[1,2,\textdaggerdbl]{Chenyan Wang}
\author[2,*]{Zhifeng Gao}
\author[1,*]{Xiaohui Yang}
\affil[1]{State Key Laboratory of Physical Chemistry of Solid Surface,\protect\\
College of Chemistry and Chemical Engineering, Xiamen University, Xiamen, China}
\affil[2]{DP Technology, Beijing, China}
\newcommand{\preprintAuthorNotes}{%
  \preprintSup{\textdaggerdbl}Work done during internships at DP Technology.\\[2pt]
  \preprintSup{*}Corresponding authors.}